\documentclass[letterpaper]{article} 
\usepackage{aaai2026}  
\usepackage{times}  
\usepackage{helvet}  
\usepackage{courier}  
\usepackage[hyphens]{url}  
\usepackage{graphicx} 
\usepackage{natbib}  
\usepackage{caption} 
\usepackage{algorithm}
\usepackage{algorithmic}

\usepackage{amsmath}

\usepackage{newfloat}
\usepackage{listings}
\usepackage{amsthm}
\usepackage{subfiles}
\usepackage[most]{tcolorbox} 
\usepackage{amssymb}

\definecolor{boxbluebg}{RGB}{232,240,252}
\definecolor{boxblueframe}{RGB}{116, 163, 239}
\definecolor{boxyellowframe}{RGB}{253,191,39}
\definecolor{boxyellow}{RGB}{254,247,227}
\definecolor{boxredframe}{RGB}{241,152,147}
\definecolor{boxred}{RGB}{251,233,231}

\newtcolorbox{myboxA}{
  colback=blue!5!white,
  colframe=blue!75!black,
  title=Theorem,
  fonttitle=\bfseries,
  boxrule=0.8mm,
  arc=1mm,
  breakable
}

\newtcolorbox{defbox}{
  colback=boxyellow,
  colframe=boxyellowframe,
  coltitle=white,
  boxrule=1pt,
  arc=1pt,
  left=6pt,
  right=6pt,
  top=6pt,
  bottom=6pt,
  enhanced,
  breakable
}

\newtcolorbox{thmbox}{
  colback=boxred,
  colframe=boxredframe,
  coltitle=white,
  boxrule=1pt,
  arc=1pt,
  left=6pt,
  right=6pt,
  top=6pt,
  bottom=6pt,
  enhanced,
  breakable
}

\newtcolorbox{sketch}{
  colback=red!5!white,
  colframe=red!75!black,
  title=Proof Sketch,
  fonttitle=\bfseries,
  boxrule=0.8mm,
  arc=0mm,
  breakable
}

\usepackage{newfloat}
\usepackage{listings}
\DeclareCaptionStyle{ruled}{labelfont=normalfont,labelsep=colon,strut=off} 
\floatstyle{ruled}
\newfloat{listing}{tb}{lst}{}
\floatname{listing}{Listing}
\title{The Alignment Game: A Theory of Long-Horizon Alignment \\ Through Recursive Curation}
\author{
    Ali Falahati\textsuperscript{\rm 1}, Mohammad Mohammadi Amiri\textsuperscript{\rm 2}, Kate Larson\textsuperscript{\rm 1}, Lukasz Golab\textsuperscript{\rm 1}
    }
\affiliations{
    \textsuperscript{\rm 1} University of Waterloo\\
        \textsuperscript{\rm 2} Rensselaer Polytechnic Institute\\

    afalahati@uwaterloo.ca
}

\usepackage{bibentry}

\begin{document}

\newtheorem{theorem}{Theorem}
\newtheorem{corollary}[theorem]{Corollary} 

\newtheorem{definition}{Definition}
\newtheorem{lemma}{Lemma}
\newtheorem{proposition}[lemma]{Proposition} 
\newtheorem{assumption}{Assumption}
\newtheorem{remark}{Remark}

\newcommand{\R}{\mathbb{R}}
\newcommand{\X}{\mathcal{X}}
\newcommand{\Z}{\mathcal{Z}}
\newcommand{\Prob}{\mathcal{P}}
\newcommand{\e}{\mathrm{e}}
\newcommand{\mech}{recursive BT-based curation }
\newcommand{\fix}{\marginpar{FIX}}
\newcommand{\new}{\marginpar{NEW}}

\maketitle

\begin{abstract}
In self-consuming generative models that train on their own outputs, alignment with user preferences becomes a recursive rather than one-time process.  We provide the first formal foundation for analyzing the long-term effects of such recursive retraining on alignment.  Under a two-stage curation mechanism based on the Bradley–Terry (BT) model, we model alignment as an interaction between two factions: the \emph{Model Owner}, who filters which outputs should be learned by the model, and the \emph{Public User}, who determines which outputs are ultimately shared and retained through interactions with the model. Our analysis reveals three structural convergence regimes depending on the degree of preference alignment: consensus collapse, compromise on shared optima, and asymmetric refinement. We prove a fundamental impossibility theorem: no recursive BT-based curation mechanism can simultaneously preserve diversity, ensure symmetric influence, and eliminate dependence on initialization. Framing the process as dynamic social choice, we show that alignment is not a static goal but an evolving equilibrium, shaped both by power asymmetries and path dependence.

\end{abstract}

\begin{links}
    \link{Extended version}{https://github.com/Mortrest/Alignment-Game/}
\end{links}

\section{Introduction}
\label{sec:intro}

Reinforcement Learning from Human Feedback (RLHF) \citep{ouyang2022training} has become the de facto method for aligning large language models with human preferences. Its appeal lies in a simple loop: broadcast model outputs to annotators, collect pairwise preferences, and update the policy through reward modeling. Among the many instantiations of this loop, the \emph{Bradley-Terry (BT) comparison model} \citep{bradley1952rank} is widely used, powering early alignment successes such as InstructGPT and many subsequent fine-tuning pipelines. At the same time, BT has been criticized for strong independence assumptions of annotating samples, a tendency to reward extremely probable rather than diverse outputs, and vulnerability to noisy feedback \citep{ge2024axioms,xiao2024algorithmic}.

Most analyses of BT ask whether a single round of collecting human preferences yields an aligned model. In practice, however, modern generative systems are updated recursively: synthetic outputs are added to the corpus, new models are trained on these added data, and the cycle repeats across model generations. Recent work shows that this self-consuming regime can drift away from human values or collapse onto degenerate equilibria \citep{ferbach2024self,shumailov2023curse,gerstgrasser2024model,alemohammad2023self}. A long-horizon view demands a formalism that tracks how preference aggregation compounds across successive rounds.

To fill this gap, we study the long-term alignment dynamics under recursive retraining. We focus on the simplest yet already rich scenario with two factions: the \textbf{Model Owner}, representing the incentives of developers or platform providers, and the \textbf{Public}, representing the aggregated preferences of society interacting with the model. At each iteration, the Owner curates samples via a mechanism based on BT, the model is trained on the selected outputs, the Public interacts with the released model and preserves a subset of model outputs through various actions such as upvoting, sharing, and other forms of engagement, and these data flow into the next training set. Our central question is: \emph{if this BT\ curation loop continues indefinitely, what distribution of content survives in the limit}?


We model this loop as a dynamic social choice mechanism, which allows us to reason about fairness, incentive compatibility, and power asymmetries in a multi-agent setting. Our analysis is structured around three core alignment scenarios, each motivated by real-world tensions. First, we explore Perfect Alignment, where the Owner and Public preferences are identical. This represents the idealized "fully-aligned" goal that many systems implicitly strive for, and analyzing it is critical to understanding the consequences of perfect agreement. Second, we examine Partial Alignment, the most realistic scenario where preferences overlap but do not perfectly coincide: for instance, a platform's commercial goals overlapping with, but not fully capturing, public interests. This case allows us to study the dynamics of compromise. Finally, we analyze Disjoint Alignment, which models a critical conflict where developer and user values are in direct opposition. This scenario reveals the system's inherent power dynamics and who ultimately controls the output when values diverge. By understanding the long-run trajectory of each regime, we can expose the structural trade-offs between diversity, fairness, and stability, establishing a foundation for alternative mechanisms and richer multi-agent systems.

In summary, our contributions are threefold: (i) a formal model of recursive BT‑based RLHF involving a model owner and a public user; (ii) a complete characterization of its long‑run behavior, including an impossibility result that generalizes classic social choice tensions to the dynamic setting; and (iii) empirical evidence that these theoretical dynamics surface in practice. This work aims to lay the groundwork for a broader theory of long-term alignment, extending beyond single-step fine-tuning to account for the recursive and evolutionary nature of generative models.

\section{Related Work}
Generative AI systems face fundamental challenges arising from recursive training dynamics and the cumulative effects of self-generated data. \citep{shumailov2023curse} and \citep{ferbach2024self} demonstrated that training models on their own generated content leads to irreversible ``model collapse,'' where output diversity diminishes through recursive iterations. These limitations are compounded by recursive training dynamics, as \citep{dohmatob2024model} demonstrates clear crossover points between stable and collapse regimes~\citep{xu2025probabilistic}. \citep{gerstgrasser2024model} showed that data accumulation strategies can prevent collapse while replacement accelerates it. 

The multi-stakeholder dimension of RLHF has been explored by \citep{tewolde2024social}, who propose Reinforcement Learning from Collective Human Feedback using social choice theory, while \citep{mishra2023ai} proves that universal AI alignment using RLHF is impossible under democratic constraints. The BT model, widely used in preference-based reward modeling, faces significant limitations in multi-stakeholder scenarios. \citep{sun2024rethinking} provides the first theoretical critique, arguing these models are unnecessary for downstream optimization and proposing classification-based alternatives focused on ``order consistency.'' \citep{zhang2024beyond} demonstrate that BT models ``fall short in expressiveness, particularly in addressing intransitive preferences'' that arise naturally in multi-stakeholder scenarios. 

\citep{wu2022diagnostic} shows that these models violate the independence assumptions when multiple stakeholders are involved, leading to systematic marginalization of the preferences of minorities. \citep{eckersley2019impossibility} demonstrates that Arrow's impossibility theorem applies to AI alignment, showing that no satisfactory method exists to aggregate multiple human preferences without violating fairness criteria. \citep{qiu2024representative} extends this with Arrow-like impossibility theorems for representative social choice settings. The convergence of theoretical impossibility results with empirical evidence of model collapse establishes the foundation for understanding AI alignment as an inherently conflicted game between competing stakeholder interests. Additional insights from social choice theory highlight its role in guiding AI alignment, particularly in handling diverse human feedback and avoiding preference aggregation pitfalls \citep{conitzer2024social}.

\section{Problem Definition}
\label{sec:problem-setup}

We study iterative retraining of generative models, where synthetic outputs recursively influence future training data through a two-agent curation process. This analysis captures the alignment challenges that emerge when two stakeholders jointly shape the evolution of AI systems.

\subsection{Model Components}

Consider a generative modeling system with the following components:
\begin{itemize}
    \item \textbf{State Space:} A compact metric space $(\mathcal{X}, d)$ representing the content domain.
    \item \textbf{Agents:} Two curators, the \emph{Owner} (model developer) and the \emph{Public} (user community), with continuous reward functions $r_O, r_P: \mathcal{X} \rightarrow \mathbb{R}$ encoding their respective preferences.
    \item \textbf{Data Evolution:} A sequence of public datasets $\{\mathcal{D}_t\}_{t \geq 1}$ where $\mathcal{D}_t \subset \mathcal{X}$.
    \item \textbf{Model Sequence:} Generative models $\{\mathcal{M}_t\}_{t \geq 1}$ with output distributions $p_t \in \mathcal{P}(\mathcal{X})$ where $\mathcal{P}(\mathcal{X})$ denote the set of all Borel probability measures on $\mathcal{X}$.
    \item \textbf{Sampling Pool:} We define a \emph{pool} of size $K$ as a collection of i.i.d.\ samples
    $
    \{x_1,\dots,x_K\} \sim p.
    $
    The integer $K$ is the \emph{pool size} and represents the number of alternatives available for pairwise comparison under BT.
    
\end{itemize}

\begin{defbox}
\begin{definition}
For a probability measure $p$ on $\mathcal{X}$, a pool size $K \geq 2$, and a reward function $r$, define the BT weight as \citep{ferbach2024self}:
\begin{align}
\label{bt-eq}
H^{p}_{K, r}(x) := \mathbb{E}_{Y_1,\ldots,Y_{K-1} \sim p}\left[\frac{K\, e^{r(x)}}{e^{r(x)} + \sum_{j=1}^{K-1} e^{r(Y_j)}}\right]. \nonumber
\end{align}
\end{definition}
\end{defbox}

\begin{tcolorbox}[
  colback=boxbluebg,
  colframe=blue!75!black,
  title={\raisebox{-0.1pt}{\textbf{Alignment Game Framework}}},
  coltitle=white,
  boxrule=1pt,
  arc=1pt,
  left=6pt,
  right=6pt,
  top=6pt,
  bottom=6pt,
  enhanced,
  breakable
]

The system recursively evolves as follows. Start with initial dataset $\mathcal{D}_1$ and initial model distribution $p_0$.  

\noindent \textbf{Recursive loop (for $t = 1, 2, \ldots$):}

\begin{enumerate}

    \item[\textbf{(1)}] \textbf{Owner Curation:} At iteration $t$, the Owner samples a pool $\{x_1,\ldots,x_K\}\sim p_t$ \emph{(the pool is a subset of $\mathcal{D}_t$, i.e., $\{x_i\}\!\sim\! q_t$ with $q_t \approx p_t$)} and selects outputs via BT selection with reward $r_O$, yielding:
    \begin{align}
        \tilde{p}_t(x) = p_t(x) \cdot H_{K, r_O}^{p_t}(x)
    \end{align}

    \item[\textbf{(2)}] \textbf{Model Update:} Train $\mathcal{M}_{t+1}$ on data drawn from $\tilde{p}_t$, producing the updated model distribution:
    \begin{align}
        p_{t+1}(x) \approx \tilde{p}_t(x)
    \end{align}

    \item[\textbf{(3)}] \textbf{Public Curation:} The Public samples a pool $\{\hat{x}_1, \ldots, \hat{x}_M\} \sim p_{t+1}$ and applies BT selection with reward $r_P$, yielding:
    \begin{align}
        \hat{p}_t(x) = p_{t+1}(x) \cdot H_{M, r_P}^{p_{t+1}}(x)
    \end{align}

    \item[\textbf{(4)}] \textbf{Dataset Evolution:} Update $\mathcal{D}_{t+1} = \mathcal{D}_t \cup \mathcal{O}_t^*$ where $\mathcal{O}_t^* \sim \hat{p}_t$.
\end{enumerate}
\end{tcolorbox}

\subsection{Alignment Regimes}

The interaction between curator preferences determines the system's long-term behavior. Let $A_O = \arg\max_{x \in \mathcal{X}} r_O(x)$ and $A_P = \arg\max_{x \in \mathcal{X}} r_P(x)$ denote the optimal sets for each curator.

\begin{defbox}
\begin{definition}
We categorize value alignment between the Owner and the Public as:
\begin{itemize}
    \item \textbf{Perfect Alignment:} $A_O = A_P$ (complete agreement)
    \item \textbf{Partial Alignment:} $A_O \cap A_P \neq \emptyset$ with $A_O \neq A_P$ (overlapping preferences)
    \item \textbf{Disjoint Alignment:} $A_O \cap A_P = \emptyset$ (conflicting preferences)
\end{itemize}
\end{definition}
\end{defbox}

\subsection{Assumptions}
\noindent Our analysis relies on the following assumptions:

\begin{enumerate}
    \item The recursive curation mechanism is explicitly defined by the BT model. The convergence properties and the impossibility theorem are direct consequences of the mathematical properties of this specific pairwise comparison model.

    \item We analyze the idealized dynamic $p_{t+1}(x) = \tilde{p}_{t}(x)$, where the model distribution at the next step perfectly matches the target distribution from the current step's curation. This abstracts away optimization error, noise, or catastrophic forgetting from practical training.

    \item The reward functions $r_O: \mathcal{X} \to \mathbb{R}$ and $r_P: \mathcal{X} \to \mathbb{R}$ are assumed to be fixed and continuous over a compact metric space $\mathcal{X}$. This ensures the optimal sets $A_O$ and $A_P$ are well-defined and non-empty.

\end{enumerate}

\noindent Throughout the analysis, we consider open neighborhoods around maximizer sets. For any set \( A \subset \mathcal{X} \) and radius \( \eta > 0 \), 
\begin{align}
B_\eta(A) := \{ x \in \mathcal{X} : \inf_{y \in A} d(x, y) < \eta \}.
\end{align}
where $d : \mathcal{X} \times \mathcal{X} \to \mathbb{R}_{\ge 0}$ is the metric on $\mathcal{X}$.

\subsection{Core Recursive Alignment Challenges}
\noindent We assume as $t \to \infty$, the public dataset becomes dominated by curated synthetic data:
\begin{align}
    \lim_{t \to \infty} \frac{|\mathcal{D}_1|}{|\mathcal{D}_t|} = 0.
\end{align}
This \emph{self-consuming} regime, where models train predominantly on their own filtered outputs, is a critical challenge for maintaining alignment with diverse human values while avoiding mode collapse or value lock-in. This setting enables us to investigate fundamental questions about recursive alignment:
\begin{itemize}
    \item \textbf{Convergence:} Under what conditions does the system converge to a stable distribution? Does it collapse to point masses or preserve diversity?
    \item \textbf{Influence Dynamics:} How do the sequential curation mechanisms affect the relative influence of each curator?
    \item \textbf{Alignment Impact:} How does the degree of preference alignment between curators shape the evolution and final state of the generative model?
\end{itemize}

\section{Theoretical Results}

We now analyze the long-term behavior of our two-stage curation mechanism under varying degrees of alignment between curators\footnote{Proofs are in Section A of Appendix in the extended version.}. We begin with the idealized case where both the Owner and the Public share identical preferences.

\subsection{Perfect Alignment: The Consensus Trap}

\begin{theorem}
\label{thm:perfect-alignment}
Let the Owner and Public have perfectly aligned preferences: $A_O = A_P =: A_\star \neq \varnothing$ and $\{p_t\}_{t \geq 0}$ be the sequence of distributions generated by this mechanism. Then, the system reaches complete consensus: for any $\eta > 0$, there exist constants $C, c > 0$ such that $p_t(\mathcal{X} \setminus B_\eta(A_\star)) \leq Ce^{-ct}$ for all $t \geq 0$. Moreover, $p_t$ converges weakly to the renormalized initial distribution on $A_\star$:
\[
p_\infty(x) = \frac{p_0(x)}{\int_{A_\star} p_0(z) \, dz} \mathbf{1}_{A_\star}(x).
\]
\end{theorem}

\noindent Our first result uncovers that greater alignment does not preserve diversity, but speeds up its collapse. When preferences are perfectly aligned, the model concentrates on the shared maximizers, leading to a sharply reduced outcome space. Perfect alignment produces an impoverished limiting distribution. The exponential decay rate $e^{-ct}$ reflects the speed of this convergence, highlighting how quickly diversity is lost under alignment. In the limit, the only remaining variation arises from the initial distribution $p_0$ restricted to $A_\star$.

\begin{corollary}[Mode Collapse for Unique Maximizers]
\label{cor:mode-collapse}
When both curators agree on a unique optimal point $A_O = A_P = \{x^\star\}$, system undergoes mode collapse: $p_t \to \delta_{x^\star}$, where $\delta_{x}$ represents the Dirac delta function at $x$.  
\end{corollary}

\noindent This corollary represents the most extreme form of homogenization and highlights a fundamental risk in alignment-driven curation: as agreement increases, the support of the distribution contracts. 
Thus, systems designed to maximize agreement between developers and users may unintentionally collapse into echo chambers.
This consensus trap suggests that some degree of preference misalignment may be necessary to sustain a healthy generative ecosystem. Our result complements prior findings such as \citep{ferbach2024self}, which show that recursive training on curated outputs can lead to long-term degeneracy. Thus, before pursuing perfect agreement, we should ask whether we are engineering an echo chamber that eliminates diversity.

\subsection{Partial Alignment: The Compromise Equilibrium}

Next, we consider the case where curators share some common ground while maintaining distinct preferences, a setting that better reflects the relationship between model owners and diverse user communities.

\begin{theorem}
\label{thm:partial-alignment}
Suppose the curators have overlapping but distinct preferences, with shared optima $A_{\text{shared}} := A_O \cap A_P \neq \varnothing$ while $A_O \neq A_P$. Under the two-stage curation mechanism, only the intersection survives: the mass outside $B_\eta(A_{\text{shared}})$ decays exponentially as $p_t(\mathcal{X} \setminus B_\eta(A_{\text{shared}})) \leq Ce^{-ct}$. The limiting distribution $p_\infty$ concentrates entirely on $A_{\text{shared}}$ with density proportional to the initial distribution:
\[
p_\infty(x) = \frac{p_0(x)}{\int_{A_{\text{shared}}} p_0(z)dz} \mathbf{1}_{A_{\text{shared}}}(x).
\]
\end{theorem}

\noindent Our findings reveal a key dynamic of partial alignment: the recursive process filters out content valued exclusively by one curator, preserving only what lies in the intersection. This “lowest common denominator” effect eliminates curator-specific signals, including potentially novel or specialized contributions. Although diversity is maintained within $A_{\text{shared}}$, it remains limited to areas of mutual agreement.
For AI systems, this suggests that iterative training with multiple stakeholders may progressively reduce the breadth of the model to only the jointly endorsed features.

\subsection{Disjoint Alignment: Asymmetric Power Dynamics}

The most adversarial situation arises when curators hold entirely disjoint preferences. In such cases, the question becomes: who ultimately defines the output space of the generative model when both sides embody incompatible values?

\begin{theorem}
\label{thm:disjoint}
When curator preferences are completely disjoint ($A_O \cap A_P = \varnothing$), the Owner determines the support while the Public refines within it. Define the Public's preferred subset within Owner optima as $A_{P|O} := \arg\max_{x \in A_O} r_P(x)$. The system exhibits two-stage exponential suppression.  First, content outside the Owner's optima vanishes with $p_t(\mathcal{X} \setminus B_\eta(A_O)) \leq C_1e^{-c_1t}$, then within $A_O$, mass concentrates on the Public's preferred subset with $p_t(A_O \setminus B_\eta(A_{P|O})) \leq C_2e^{-c_2t}$. The limiting distribution is
\[
p_\infty(x) = \frac{p_0(x)}{\int_{A_{P|O}} p_0(z)dz} \mathbf{1}_{A_{P|O}}(x).
\]
\end{theorem}

\begin{figure*}[t]
\centering
\includegraphics[width=0.71\textwidth]{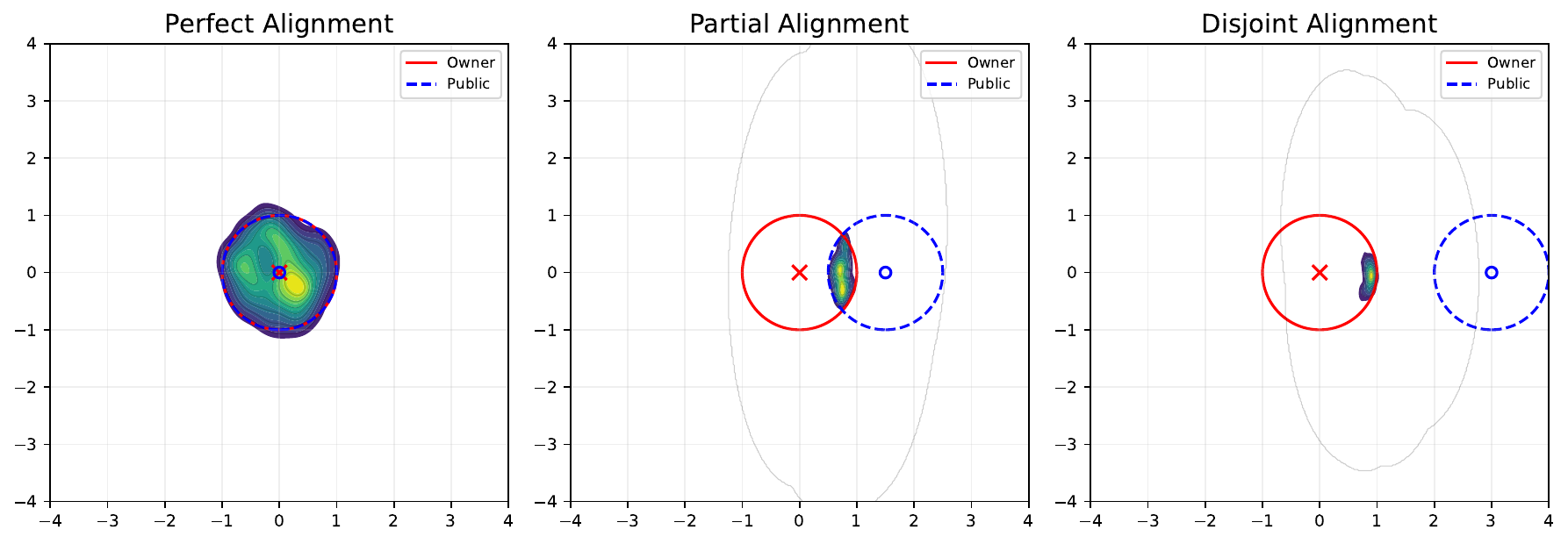}
\caption{\label{fig:set_based} KDE plots showing point distributions in three alignment scenarios: perfect alignment (same circles), partial alignment (overlapping circles), and disjoint alignment (non-overlapping circles). Red circles indicate owner preferred regions, blue dashed circles show public preferred regions.
}
\end{figure*}

\noindent This result exposes the inherent power asymmetry in the recursive alignment game. The Owner's first-mover advantage allows it to determine the feasible region, while the Public can only optimize within these constraints. This ``best of the worst'' dynamic mirrors real-world scenarios where users must select from options pre-filtered by platform algorithms. The Owner's preferences shape the system quickly, while the Public's influence manifests gradually as a refinement. This temporal divide reveals a deeper governance dilemma: once early alignment choices solidify into norms, user feedback becomes more about adaptation than agency.

\subsection{The Fundamental Impossibility Result}

After exploring the range of alignment scenarios, we turn to a deeper limitation of recursive BT-based curation. Even under ideal conditions, certain properties that appear simultaneously desirable cannot all be achieved at once.

\begin{theorem}
\label{thm:impossibility}
For any non-trivial preference misalignment $(A_O \neq A_P)$, it is impossible to simultaneously satisfy:
\begin{enumerate}
    \item[(1)] Full Coverage: 
    \[
    \liminf_{t \to \infty} p_t(A_O \setminus A_P) > 0 \quad \text{and} \quad \liminf_{t \to \infty} p_t(A_P \setminus A_O) > 0
    \]
    
    \item[(2)] Symmetric Influence: There exists a permutation-invariant functional $\Phi$ such that
    \[
    p_\infty = \Phi(r_O, r_P, p_0) = \Phi(r_P, r_O, p_0)
    \]
    
    \item[(3)] Initial Distribution Independence: For any two initial distributions $p_0, q_0 \in \mathcal{P}(\mathcal{X})$ with $\operatorname{supp}(p_0) = \operatorname{supp}(q_0) = \mathcal{X}$,
    \[
    p_\infty^{(p_0)} = p_\infty^{(q_0)}
    \]
\end{enumerate}
\end{theorem}

\begin{figure*}[t]
\centering
\includegraphics[width=0.67\textwidth]{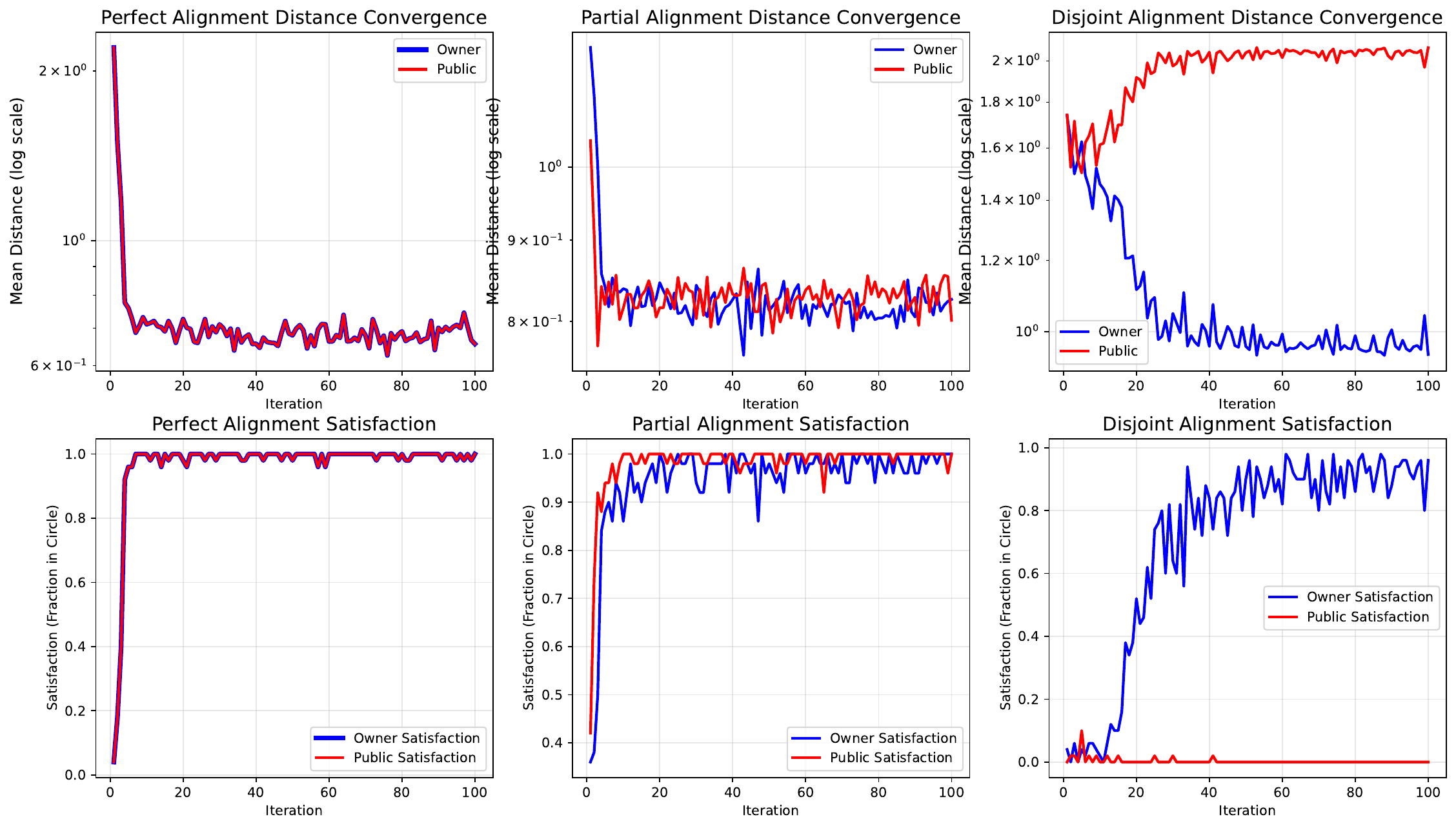}
\caption{\label{fig:convg} (Top) Convergence of the mean distance to the owner and public centers over iterations for three alignment scenarios: perfect alignment, partial alignment, and disjoint alignment.(Bottom) The fraction of points within the owner and public preferred regions (“satisfaction”) as the alignment process progresses.
}
\end{figure*}

\noindent This impossibility result states that no \mech mechanism can simultaneously maintain content diversity across disagreement regions (Full Coverage), treat both curators equally (Symmetric Influence), and produce outcomes independent of initial conditions (Initial Distribution Independence).
Each of these properties captures a fundamental value tension in alignment systems. Full Coverage represents the epistemic goal of preserving the full spectrum of ideas, ensuring that minority or dissenting content is not prematurely filtered out. Symmetric Influence encodes fairness: both curators should exert comparable control over the generative process, preventing dominance by either side. Initial Distribution Independence embodies stability, guaranteeing that the system’s long-term behavior does not hinge on arbitrary starting points or early biases.

The theorem’s implications extend beyond our specific model: it formalizes the intuition that these goals pull in incompatible directions. The tradeoffs observed in \mech systems are therefore not artifacts of design but mathematical necessities. Designers must decide which virtue to compromise: accepting homogenization, embracing asymmetry, or tolerating historical lock-in.
\begin{thmbox}
\begin{remark}
\label{rem:shrinking}
Across all alignment regimes, the support of the limiting distribution shrinks as follows:
\begin{itemize}
    \item $\operatorname{supp}(p_\infty) \subseteq A_O \cap A_P$ when the intersection is non-empty
    \item $\operatorname{supp}(p_\infty) \subseteq A_{P|O} \subseteq A_O$ when $A_O \cap A_P = \emptyset$
\end{itemize}
\end{remark}
\end{thmbox}

\noindent This demonstrates that recursive BT-based curation inevitably reduces diversity, concentrating mass on increasingly narrow regions of agreement.
\noindent These structural constraints raise a deeper question: what kind of decision-making process is recursive curation, and how should we reason about its limitations? To answer this, we now reinterpret the \mech mechanism using social choice theory.

\section{The Curation Mechanism as Social Choice}
The \mech process can be viewed as a form of collective decision-making, where two agents jointly shape the long-term distribution of model outputs. To examine its properties, fairness, efficiency, and the preservation of diversity, we draw from social choice theory and mechanism design. We formalize \mech as a dynamic preference aggregation mechanism in which the Owner and the Public iteratively express preferences over a shared alternative space $\mathcal{X}$ through a two-stage influence process analogous to voting.

\begin{defbox}
\begin{definition}
A dynamic social choice mechanism for the alignment game consists of:
\begin{enumerate}
    \item[\textbf{(1)}] A set of alternatives $\mathcal{X}$ (the content space).
    \item[\textbf{(2)}] Two agents $\mathcal{N} = \{O, P\}$ (Owner and Public).
    \item[\textbf{(3)}] Preferences, captured by reward functions $r_i: \mathcal{X} \to \mathbb{R}$, for $i \in \mathcal{N}$.
    \item[\textbf{(4)}] A social choice correspondence $\mathcal{F}: \mathcal{R}^2 \times \mathcal{P}(\mathcal{X}) \to \mathcal{P}(\mathcal{X})$ that maps preference profiles and current distributions to updated distributions.
    \item[\textbf{(5)}] A limit social choice function $f_\infty: \mathcal{R}^2 \to \mathcal{P}(\mathcal{X})$ representing the long-run outcome.
\end{enumerate}
\end{definition}
\end{defbox}
\noindent The mechanism implements a sequential decision process in which agents express preferences via pairwise (binary) comparisons over alternatives. The Owner selects an alternative \( x \in \mathcal{X} \) first, based on their reward function, and the Public responds by evaluating this choice using their own reward function. The final distribution over alternatives is updated based on these sequential comparisons, reflecting the asymmetric influence of each agent.

\subsection{Strategic Voting and Incentive Compatibility}

In the social choice literature, a central property is strategyproofness: agents should not have an incentive to misrepresent their preferences. 

\begin{theorem}
\label{thm:strategy}
For any reported pair $(r'_O,r'_P)$, let $p_\infty(r'_O,r'_P)$ denote the weak limit of $\{p_t\}$. Define each agent’s utility by
\begin{align} 
U_O(r'_O,r'_P):=\mathbb E_{x\sim p_\infty(r'_O,r'_P)}[\,r_O(x)\,], \nonumber \\
U_P(r'_O,r'_P):=\mathbb E_{x\sim p_\infty(r'_O,r'_P)}[\,r_P(x)\,]. \nonumber
\end{align}
Then, truthful reporting is weakly dominant for both agents:
\begin{align}
    U_O(r_O,r'_P)\;\ge\;U_O(\hat r_O,r'_P), \quad
U_P(r'_O,r_P)\;\ge\;U_P(r'_O,\hat r_P) \nonumber
\end{align}
for all $(\hat r_O,\hat r_P)$, and for all $(r'_O,r'_P)$. In particular, $(r'_O,r'_P)=(r_O,r_P)$ is a Nash equilibrium.
\end{theorem}

\noindent The theorem ensures truthful elicitation: for both agents, truthful reporting weakly dominates any misreport, regardless of the other report. Manipulating reports is therefore ineffective; influence instead comes from the dynamics. The result addresses incentives over reports only, and the broader questions of fairness, symmetric influence, and coverage are taken up in the next theorem, which formalizes their mutual incompatibility.

\subsection{Fairness and Representation in Dynamic Voting}

Beyond strategyproofness, social choice theory provides principled ways of thinking about fairness criteria, such as equal treatment of agents. We formalize this notion through the concept of \emph{influence parity}.

\begin{theorem}
\label{thm:impossibility-social}
No asymmetric sequential preference aggregation mechanism can simultaneously satisfy:
\begin{enumerate}
\item[(EQ)] Influence parity: the outcome is order symmetric, that is
\[
p_\infty^{\mathrm{OF}}(r_O,r_P)\;=\;p_\infty^{\mathrm{PF}}(r_O,r_P).
\]
Here $p_\infty^{\mathrm{OF}}$ is the limiting distribution when the Owner acts first, and $p_\infty^{\mathrm{PF}}$ when the Public acts first. 
\item[(PO)] Pareto optimality: there is no distribution $q$ on $X$ with
\[
\mathbb E_q[r_O]\ge \mathbb E_{p_\infty}[r_O],\qquad
\mathbb E_q[r_P]\ge \mathbb E_{p_\infty}[r_P],
\]
\item[(UC)] Uniqueness: for each $(r_O,r_P)$ the outcome is unique, independent of initialization and of any tie breaking.
\end{enumerate}
\end{theorem}

\noindent The sequential nature of voting creates a first-mover advantage that conflicts with equal treatment. Unlike static voting where ties can be broken symmetrically, the recursive dynamics amplify initial asymmetries, making true voter equality impossible while maintaining uniqueness.

\begin{figure*}[t]
\centering
\includegraphics[width=0.8\textwidth]
{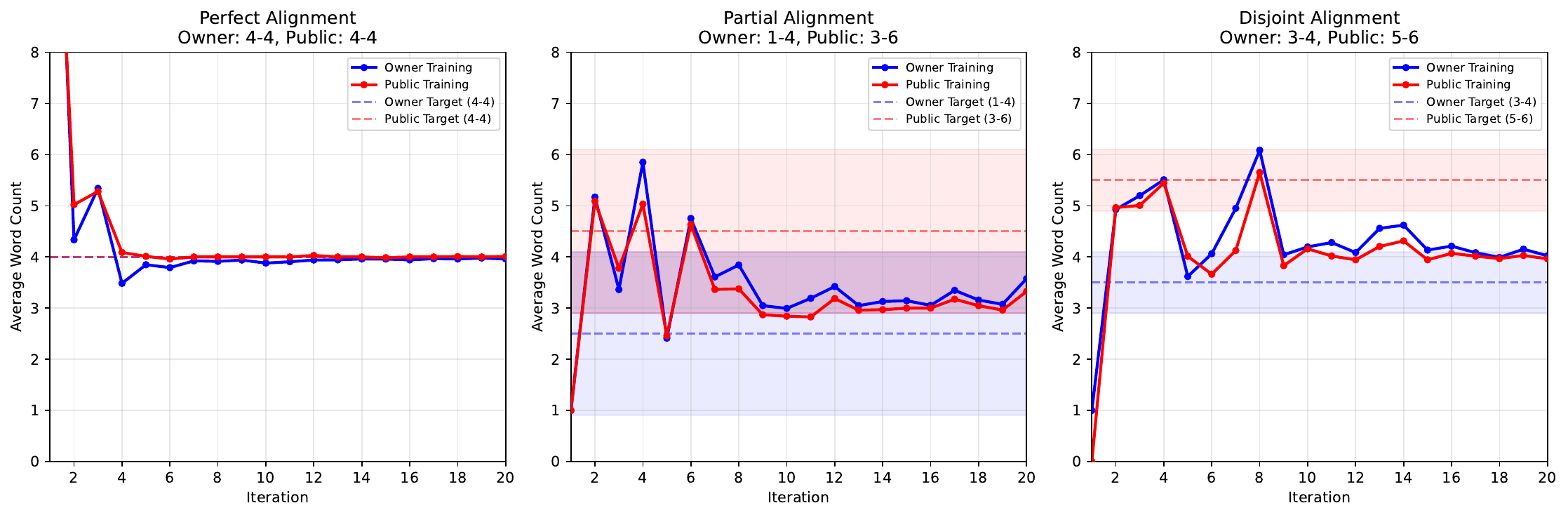}
\caption{\label{fig:text_based}Word count evolution across three alignment scenarios: perfect alignment (agents target 4 words), partial alignment (owner targets 1-3 words, public targets 3-5 words), and disjoint alignment (owner targets 3-4 words, public targets 5-6 words).}
\end{figure*}

\vspace{-0.6em}

\section{Experiments}
\label{sec:experiments}

We validate our analysis through two experimental frameworks: a synthetic alignment game that directly implements our mathematical framework, and a text-based alignment game based on a realistic language modeling setting.

\vspace{-0.6em}

\subsection{Synthetic Alignment Game}

\noindent\textbf{Experimental Setup.} We implement a synthetic alignment game in $\mathbb{R}^2$ where the Owner and Public preferences are defined by circular reward regions. The reward function for agent $i \in \{O, P\}$ with center $c_i$ and radius $r$ is:

\begin{align}
r_i(x) = \begin{cases}
1.0 & \text{if } \|x - c_i\| \leq r \\
-2.0 \cdot (\|x - c_i\| - r) & \text{if } \|x - c_i\| > r
\end{cases}
\end{align}

\noindent This creates a sharp preference boundary where points inside the circle receive maximum reward and points outside receive penalties proportional to their distance.

\vspace{0.2em}

\noindent \textbf{Alignment Scenarios.} \emph{ (i) Perfect alignment:} $c_O = c_P = (0, 0)$, $r = 1.0$, i.e., both agents prefer the same circular region; \emph{ (ii) Partial alignment:} $c_O = (0, 0)$, $c_P = (1.5, 0)$, $r = 1.0$, i.e., overlapping circles with shared optima; \emph{ (iii) Disjoint alignment:} $c_O = (0, 0)$, $c_P = (3, 0)$, $r = 1.0$, i.e., nonoverlapping circles with no shared optima.  

\vspace{0.2em}

\noindent \textbf{Experimental Parameters.} Each experiment runs for 100 iterations with the following parameters: (i) Initial dataset: 1000 points sampled uniformly from $[-5, 5] \times [-5, 5]$; (ii) Owner curation: Select 100 points using the BT mechanism with temperature $\tau = 0.5$; (iii) Generation: Train a Gaussian Mixture Model (GMM)~\citep{bishop2006pattern} on curated data, generate 200 new samples; (iv) Public curation: Select 50 points from generated samples using the BT mechanism; (v) Dataset update: Add public-curated samples to the training.

\vspace{0.2em}

\noindent \textbf{Results and Analysis.} Figure~\eqref{fig:set_based} (top) shows the final distribution of points after 100 iterations for each alignment scenario. The KDE plots reveal convergence patterns that validate our theoretical predictions. \emph{Perfect Alignment:} The system converges to a concentrated distribution within the shared optimal region, with exponential suppression outside the circle. The final distribution shows high density within the preferred region and negligible mass elsewhere, confirming Theorem~\ref{thm:perfect-alignment}. \emph{Partial Alignment:} The system concentrates on the intersection of the two circles, preserving diversity only where preferences align. Points outside the intersection are exponentially suppressed, while the shared region maintains substantial density. This validates Theorem~\ref{thm:partial-alignment}. 

\emph{Disjoint Alignment:} The system converges to the Public's preferred subset within the Owner's optimal region. The Owner's first-mover advantage determines the support, while the Public refines the distribution within that support. This confirms Theorem~\ref{thm:disjoint}. Figure~\ref{fig:convg} (top) tracks the mean distance to each agent's preferred center over iterations. All scenarios show exponential convergence, with the rate depending on the degree of alignment. Perfect alignment achieves the fastest convergence, while disjoint alignment shows a two-stage process, first converging to the Owner's region, then refining within it. Figure~\ref{fig:convg} (bottom) shows the fraction of points satisfying each agent's preferences ("satisfaction rate"). In perfect alignment, both agents achieve near-universal satisfaction. In partial alignment, satisfaction is limited to the intersection region. In disjoint alignment, the Owner maintains high satisfaction, while the Public's satisfaction is constrained by the Owner's preferences.

\vspace{-0.6em}

\subsection{Text-Based Alignment Game}

\noindent\textbf{Experimental Setup.} We implement a realistic text generation scenario using GPT-2 models~\citep{radford2019language} where the Owner and Public have different preferences for response length. The system operates on the WikiText-2~\citep{merity2016pointer} dataset, with agents preferring different word count ranges for generated responses.

\vspace{0.2em}

\noindent\textbf{Alignment Scenarios.} We test three text-based alignment scenarios: \emph{(i) Perfect Alignment:} Both agents prefer exactly 4 words (Owner: 4-4, Public: 4-4); \emph{(ii) Partial Alignment:} Owner prefers 2-4 words, Public prefers 4-6 words (overlapping range); \emph{Disjoint Alignment:} Owner prefers 1-3 words, Public prefers 5-6 words (non-overlapping ranges).

\vspace{0.2em}

\noindent\textbf{Experimental Parameters.} (i) Initial dataset: 10000 sentences from WikiText-2 filtered by word count; (ii) Owner curation: Select 1000 sentences using the BT mechanism (iii) Generation: Fine-tune the model for 2 epochs, generate 2000 responses with temperature 0.8; (iv) Public curation: Select 1000 responses using the BT mechanism (v) Training: Learning rate $5 \times 10^{-5}$, batch size 8.

\vspace{0.2em}

\noindent\textbf{Results and Analysis.} Figure~\ref{fig:text_based} shows the evolution of average word count across iterations for all three scenarios. The results demonstrate the same convergence patterns as the synthetic experiments. \emph{Perfect Alignment:} Both training stages converge to the shared target of 4 words within 10 iterations, achieving near-perfect alignment with minimal variance. \emph{Partial Alignment:} The system converges to the intersection of preferences (around 3-4 words), with the Owner's training showing slightly lower word counts and the Public's training showing slightly higher counts, but both remaining within the overlapping range. \emph{Disjoint Alignment:} The system shows a two-stage convergence process. Initially, both training stages converge toward the Owner's preferred range (3-4 words). However, the Public's training gradually shifts toward longer responses, eventually settling around 4 words, demonstrating the Public's ability to refine within the Owner's preferred domain.
\vspace{0.2em}

\noindent\textbf{Key Findings.} Our experiments validate several theoretical predictions. All scenarios show exponential convergence to equilibrium distributions, with convergence rates inversely proportional to the degree of alignment. In all cases, the final distribution concentrates on a subset of the original support, confirming the shrinking support principle (Remark~\ref{rem:shrinking}). In disjoint alignment scenarios, the Owner’s preferences dominate the support selection, while the Public refines within that support. Partial alignment scenarios preserve diversity only within the intersection of preferred regions, with exponential suppression elsewhere. These phenomena persist across different domains (continuous vs. discrete text) and different model architectures (GMM vs. transformer).

\vspace{-0.4em}
\section{Conclusion}
\label{sec:conclusion}
Alignment is not a one-time setting. In recursive curation, even well-intended preference aggregation can reduce diversity, confer first mover power, and lock models into narrow equilibria. We identified three convergence regimes, consensus collapse, lowest common denominator compromise, and owner-led refinement, showing how small preference gaps can redirect the long-run trajectory of generative systems. 

\vspace{0.3em}
\noindent\textbf{Implications for AI alignment.}
Our findings suggest that we must reconceptualize alignment itself. First, it moves the problem from one-time preference-matching to dynamic mechanism design. The paper demonstrates that the structure of the alignment process, the sequential curation, the power dynamics, is a value-driven system that actively shapes outcomes, privileging owner-defined constraints over public refinement. Second, the impossibility theorem is not a failure but a clarification. It suggests that the goal of a single, stable, and diverse alignment may be a contradiction in terms. This should force the field to pivot from designing for consensus to designing for pluralism. The challenge is not to eliminate disagreement but to build systems that can productively manage it, treating "lowest common denominator" outcomes as a failure state of the mechanism, not an acceptable compromise. Finally, this work implies that the true alignment "meta-problem" is not just aligning a model, but aligning the recursive alignment process itself. We must design the feedback loop, the "game", to be transparent, contestable, and fair. Instead of a one-time setting, alignment becomes a continuous process of governing the system that governs the models.

\vspace{0.3em}
\noindent\textbf{Limitations and Future Work.}
This analysis is based on a model with key simplifying assumptions. First, the analysis rests on the BT model, which assumes that preferences are independent and transitive. Future work must extend this analysis to more complex preference models. Second, the two-agent "Owner-Public" game is a simplification of a complex ecosystem. The next step is to model this as an n-agent game with heterogeneous agents. Third, the model assumes static preferences. A more realistic analysis would treat preferences and model outputs as co-evolving. Future research should explore this co-evolutionary dynamic, where the outputs of one generation's model actively shape the preferences of the agents who curate the next. Finally, this paper is descriptive; it explains what will happen. The next step is normative mechanism design. Given the impossibility theorem, how can we design new curation systems that explicitly and transparently choose which property to sacrifice? This moves the alignment challenge beyond the realm of optimization and into the domain of political and social governance. The task is no longer to discover a single, mathematically "correct" alignment, but to engineer a process that is perceived as legitimate, transparent, and fair by all stakeholders.

\section{Acknowledgements}
The authors thank the Natural Sciences and Engineering Research Council of Canada, the University of Waterloo José Blakeley Scholarship
for financial support. We are also grateful to the reviewers for their valued feedback on the paper.

\bibliography{aaai2026}

\end{document}